\documentclass[runningheads]{llncs}

\usepackage{eccv}

\usepackage{eccvabbrv}

\usepackage{graphicx}
\usepackage{booktabs}
\usepackage{arydshln}
\usepackage{mathtools}
\usepackage{multirow}
\usepackage[accsupp]{axessibility}  %

\makeatletter
\renewcommand\paragraph{\@startsection{paragraph}{4}{\z@}%
	{0.7ex \@plus.2ex \@minus.2ex}%
	{-1em}%
	{\normalfont\normalsize\bfseries\maybe@addperiod}}
\newcommand{\maybe@addperiod}[1]{#1\@addpunct{.}}
\makeatother

\definecolor{winered}{RGB}{115, 0, 57} 
\makeatletter
\def\adl@drawiv#1#2#3{%
        \hskip.5\tabcolsep
        \xleaders#3{#2.5\@tempdimb #1{1}#2.5\@tempdimb}%
                #2\z@ plus1fil minus1fil\relax
        \hskip.5\tabcolsep}
\newcommand{\cdashlinelr}[1]{%
  \noalign{\vskip\aboverulesep
           \global\let\@dashdrawstore\adl@draw
           \global\let\adl@draw\adl@drawiv}
  \cdashline{#1}
  \noalign{\global\let\adl@draw\@dashdrawstore
           \vskip\belowrulesep}}
\makeatother

\usepackage{hyperref}

\usepackage{orcidlink}

\begin{document}

\title{On the Diffusibility of High-Dimensional Latents}

\author{
    Chao Feng$^{1}$ \quad
    Zhiyang Xu$^{3}$ \quad
    Bowei Chen$^{4}$ \quad
    Yuanjun Xiong$^{2}$ \quad
    Xiyao Wang$^{5}$ \\
    Jui-Hsien Wang$^{2}$ \quad
    Richard Zhang$^{2}$ \quad
    Zhe Lin$^{2}$ \quad
    Andrew Owens$^{1}$ \quad
    Yijun Li$^{2}$ \\
}

\authorrunning{C. Feng et al.}

\institute{$^{1}$Cornell University~~
$^{2}$Adobe~~
$^{3}$Virginia Tech \\
$^{4}$University of Washington~~
$^{5}$University of Maryland \\
\email{cf583@cornell.edu} \\
\vspace{1em}\url{https://cfeng16.github.io/on_the_diffusibility/}}

\maketitle

\begin{abstract}
Representation Autoencoders (RAEs) enable diffusion models to operate in the
feature spaces of pretrained visual encoders. However, many off-the-shelf encoders are
not optimized for faithful reconstruction, discarding fine-grained visual
details. As expected, finetuning these encoders for image reconstruction recovers such
details. However, perhaps counterintuitively, this procedure reduces the effective dimensionality of the
resulting representation, and the altered geometry has downstream effects on generation. Specifically, we show that using the standard velocity prediction in flow matching in this high-dimensional space requires
the model to fit orthogonal noise directions outside the low-dimensional signal
manifold, making optimization inefficient. This motivates using the clean data
parameterization ($\boldsymbol{x}_{0}$-prediction) instead, which focuses learning on the
underlying signal manifold. Across experiments with multiple
strong-reconstruction encoders, we show that $\boldsymbol{x}_{0}$-prediction
consistently improves text-to-image generation performance.

  \keywords{Text-to-image generation \and Representation autoencoder \and $\boldsymbol{x}_{0}$-prediction}
\end{abstract}

\section{Introduction}
\label{sec:intro}

Diffusion and flow-matching models~\cite{ho2020denoising,song2020score,lipman2022flow,liu2022flow} have emerged as a standard paradigm for high-fidelity visual generation, including text-to-image synthesis~\cite{saharia2022photorealistic,ramesh2022hierarchical,rombach2022high}. To improve their computational efficiency, it is common to perform generation in latent space. Early work used features produced by variational autoencoders~(VAEs)~\cite{Kingma2013AutoEncodingVB,rombach2022high}. Since these latent spaces are low dimensional and often discard perceptually important information~\cite{zheng2025diffusion,tong2026scaling}, an emerging line of work on {\em representational autoencoders}~ (RAEs)~\cite{zheng2025diffusion,tong2026scaling,singh2026improved} has aimed to perform generation in the feature space of pretrained semantic visual encoders, such as self-supervised features~\cite{oquab2023dinov2} and vision-language features~\cite{tschannen2025siglip}. %

\begin{figure}[!t]
\centering

\includegraphics[width=\linewidth]{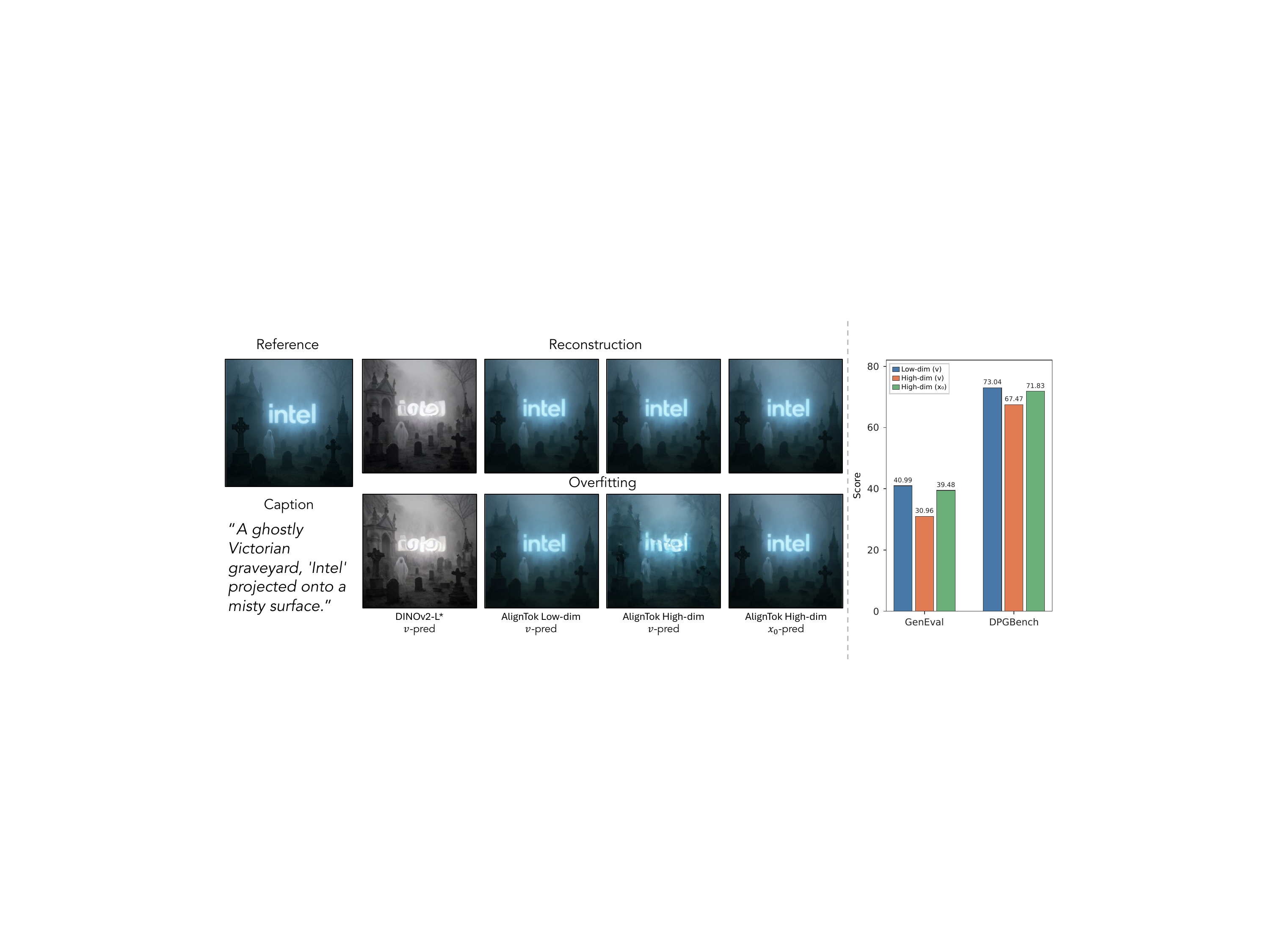}

    \begin{flushleft}
        \vspace{-3mm}
        \hspace{12mm} (a)  Reconstruction and overfitting experiment\hspace{18mm} (b) Metrics
         \vspace{-5mm}
    \end{flushleft} 

\caption{
\textbf{Reconstruction and learnability of high-dimensional representation autoencoders.}
Finetuning by reconstruction improves detail preservation but makes simple single-image overfitting more difficult under standard velocity prediction in flow matching. Clean data~($\boldsymbol{x}_0$) prediction mitigates this optimization difficulty.
(a) Frozen DINOv2-L~\cite{oquab2023dinov2} latents can be fit quickly with $\boldsymbol{v}$-prediction, but have limited reconstruction fidelity. Reconstruction-tuned DINOv2-L and MAE-RAE~\cite{he2022masked,zheng2025diffusion} produce more faithful reconstructions, yet $\boldsymbol{v}$-prediction converges slowly in a simple single-image overfitting test. In contrast, $\boldsymbol{x}_0$-prediction optimizes more efficiently in these strong-reconstruction representation spaces.
(b) This optimization advantage transfers to text-to-image generation, where $\boldsymbol{x}_0$-prediction improves GenEval~\cite{ghosh2023geneval} and DPG-Bench~\cite{hu2024ella}.
}

\vspace{-1.5em}
\label{fig:teaser}
\end{figure}

However, in contrast to the VAEs used in traditional latent diffusion models, there is no guarantee that these semantic encoders capture all of the information that is present in an image, since they are not explicitly trained for reconstruction. 
As shown in \cref{fig:teaser}, this can
lead to the loss of high-frequency information such as text, fine textures, and
small objects. %
A natural way to address this is by finetuning the semantic
encoders with an autoencoding loss~\cite{chen2025aligning,zhang2025both,tang2025unilip,
yue2025uniflow,shi2025latent}.

While these reconstruction-tuned feature spaces often improve generation quality, we find that training text-to-image diffusion models on them can be surprisingly challenging under standard formulations. %
As shown in~\cref{fig:teaser}, standard velocity prediction with flow matching~\cite{lipman2022flow,liu2022flow} becomes harder to optimize in the resulting high-dimensional feature space, converging slowly even in a simple single-image overfitting test.

\looseness=-1
Prior work addresses this issue by compressing the latents down to a lower dimensional space~\cite{zhang2025both,chen2025aligning,tang2025unilip}, but this can discard important information. Diffusion is not performed directly in the reconstruction-tuned high-dimensional feature space, leaving the potential of these representations for generation largely unexplored.

In this paper, we aim to address these optimization challenges and train image generation models directly in high-dimensional, reconstruction-tuned feature spaces.
We observe that this optimization difficulty is closely tied to the \emph{effective dimensionality}~(the minimum number of components required to explain a certain percentage of total variance) of reconstruction-tuned representations. 
Although the feature dimension is large (e.g., 768 or 1024), the learned features concentrate near a much lower-dimensional subspace, exhibiting a sharp collapse in effective dimensionality.
Under this geometry, the standard velocity target decomposes into a manifold-aligned component plus an orthogonal component that is largely uninformative about the clean representation. In high dimensions, this orthogonal term can dominate, making optimization inefficient and degrading sample quality.

Guided by this analysis, we propose a simple but effective change. Instead of predicting velocity, we take inspiration from  Li and He~\cite{li2025back} and directly predict the clean representation ($\boldsymbol{x}_0$-prediction) in the high-dimensional feature space. 
We identify and quantify a reconstruction-induced effective-dimensionality collapse in high-dimensional latent spaces, and analyze why the resulting geometry makes standard $\boldsymbol{v}$-prediction inefficient in this regime.
By focusing learning on the low-dimensional signal component and avoiding explicit regression of orthogonal noise directions, $\boldsymbol{x}_0$-prediction yields faster convergence and improved generation. Across two strong-reconstruction tokenizers (fine-tuned DINOv2-L and an MAE-based RAE~\cite{he2022masked,zheng2025diffusion}), $\boldsymbol{x}_0$-prediction consistently improves text-to-image performance on GenEval~\cite{ghosh2023geneval}, DPG-Bench~\cite{hu2024ella}, and COCO-30k FID~\cite{lin2014microsoft}, matching or surpassing the corresponding frozen semantic baselines.

\vspace{2mm}
\noindent Our main contributions are threefold:
\def\labelitemi{\textbullet}
\begin{itemize}
\item We identify and quantify an effective dimensionality collapse in reconstruction-tuned high-dimensional representations, and show that this collapse is correlated to slow convergence of standard velocity prediction.

\item We analyze why $\boldsymbol{x}_0$ prediction is better suited to this regime, showing that it mathematically bypasses the orthogonal noise components that hinder standard velocity targets in high dimensions. This allows the diffusion model to efficiently isolate and learn the underlying low-dimensional signal.

\item We demonstrate that $\boldsymbol{x}_0$-prediction consistently improves text-to-image generation in strong-reconstruction representation spaces, enabling a finetuned model to effectively surpass the generation quality of its frozen semantic baseline counterpart.

\end{itemize} 
\section{Related Work}

\subsection{Diffusion and Flow Matching}
Diffusion and flow matching models have emerged as an important paradigm for text-to-image generation. Transitioning from early U-Net-based architectures~\cite{nichol2021improved,ho2020denoising,ronneberger2015u} to modern diffusion transformers (DiT)~\cite{peebles2023scalable,ma2024sit}, these models have exhibited exceptional scalability and synthesis quality for high-fidelity generation~\cite{saharia2022photorealistic,ramesh2022hierarchical,rombach2022high,wu2025qwen,cai2025z,flux,esser2024scaling}.
Diffusion models~\cite{dhariwal2021diffusion,ho2020denoising,sohl2015deep,song2020denoising,song2020score,nichol2021improved,feng2025gps,feng2025masked} define a probability path from noise to data through a predefined noise scheduler, which specifies how signal and noise are mixed over time. Training learns to reverse this path via iterative denoising, with the network commonly parameterized to predict noise $\boldsymbol{\epsilon}$~\cite{ho2020denoising}, clean data $\boldsymbol{x}_0$~\cite{song2020denoising}, velocity $\boldsymbol{v}$~\cite{salimans2022progressive}, or the score function $\nabla_{\boldsymbol{x}_t} \log p_t(\boldsymbol{x}_t)$~\cite{song2019generative}.
Flow matching instead defines an explicit deterministic straight-line path between noise and data, and directly trains a neural network to predict the velocity field that transports samples along this path~\cite{lipman2022flow,liu2022flow}.
Under a unified probability path perspective~\cite{lipman2022flow,kingma2023understanding}, diffusion and flow matching mainly differ in the choice of path and parameterization, which in turn induces different time weightings in the training objective.

\subsection{Unifying Representation Learning and Flow Matching}
Recent work shows that strong representation learning benefits generative modeling, and existing approaches fall into two directions. The first direction aligns diffusion model features with pretrained representations during diffusion training~\cite{yu2024representation,wu2025representation,wang2025repa,singh2025matters}. For example, REPA~\cite{yu2024representation} explicitly supervises the intermediate features of diffusion models to match semantic encoder features~\cite{oquab2023dinov2,he2022masked,tschannen2025siglip}, thereby injecting structured semantic priors directly into the diffusion model. This alignment improves semantic consistency and generation quality without modifying the latent space or redesigning the overall diffusion architecture.

The second direction redesigns the latent space itself via representation autoencoders~\cite{yao2025reconstruction,chen2025masked,zheng2025vision,chen2025aligning,zheng2025diffusion}. 
Specifically, AlignTok~\cite{chen2025aligning} proposes a three-stage fine-tuning strategy for pretrained encoders, enabling them to capture high-frequency details for improved reconstruction while preserving semantic structure for better generation. However, it typically operates in lower-dimensional latent spaces, since diffusion models are difficult to train effectively in high-dimensional representation spaces.
In contrast, RAE~\cite{zheng2025diffusion} leverages pretrained encoders~\cite{oquab2023dinov2,he2022masked,tschannen2025siglip} as frozen feature extractors and demonstrates that diffusion models can operate directly on such high-dimensional semantic latents using a wider diffusion head and modified noise scheduling. However, the use of a frozen encoder limits reconstruction quality.

Our work follows the second line of work. Following AlignTok~\cite{chen2025aligning}, we extend it to high-dimensional latent spaces, which naturally improve reconstruction quality, and demonstrate that such representations can still be diffused effectively. Moreover, we find that simply adopting RAE-style designs is not sufficient to fully address the diffusibility challenges~\cite{skorokhodov2025improving,lai2026toward} in high-dimensional latent spaces, motivating a more principled treatment of this regime.

\subsection{Clean image ($\boldsymbol{x}_{0}$) prediction}

Diffusion models admit several parameterizations of the reverse process, including
predicting the reverse-process mean, the added noise $\boldsymbol{\epsilon}$, the
clean data $\boldsymbol{x}_0$, or the velocity $\boldsymbol{v}$. Early diffusion
probabilistic models learned reverse transition statistics directly~\cite{sohl2015deep}. DDPM popularized $\boldsymbol{\epsilon}$-prediction as a simple and effective
training target, while also discussing alternatives such as $\boldsymbol{x}_0$-prediction~\cite{ho2020denoising}.
Clean image prediction is also natural in image restoration, where the goal is to
recover the underlying clean signal~\cite{delbracio2023inversion,
milanfar2025denoising,xie2023diffusion}.

Recent work such as JiT~\cite{li2025back} revisits $\boldsymbol{x}_0$-prediction
and argues that it is particularly beneficial in high-dimensional settings, where
clean images lie near a low-dimensional manifold while noise targets do not. Their
analysis mainly focuses on pixel space. In contrast, we study $\boldsymbol{x}_0$-prediction
in high-dimensional representation spaces, and connect its benefit to the
effective-dimensionality collapse induced by reconstruction tuning.

\section{Method}
We first review the preliminaries of flow matching. Then, we analyze why the standard $\boldsymbol{v}$-prediction objective is difficult to optimize in high-dimensional spaces tuned for reconstruction, motivating $\boldsymbol{x}_{0}$-prediction as a simple alternative.
\subsection{Preliminaries}
\paragraph{Flow matching} In standard practice, flow matching~\cite{lipman2022flow,liu2022flow} uses linear interpolation between clean data $\boldsymbol{x}_{0} \sim p_{\text{data}}(\boldsymbol{x}) $ and randomly sampled isotropic Gaussian noise $\boldsymbol{\epsilon} \sim \mathcal{N}(0, \mathbf{I})$ as input: $\boldsymbol{z}_{t} = (1-t)\boldsymbol{x}_{0} + t\boldsymbol{\epsilon}$, where $t \in [0, 1]$ sampled from predefined time schedule. The velocity $\boldsymbol{v} = \boldsymbol{\epsilon} - \boldsymbol{x}_{0}$ is employed as the training target to supervise models. The loss function is defined as follows:
\begin{equation}
    \mathcal{L} = \mathbb{E}_{t, \boldsymbol{x}_{0}, \boldsymbol{\epsilon}} \| \boldsymbol{v}_\theta(\boldsymbol{z}_{t}, t) - \boldsymbol{v} \|^2,
\end{equation}
where $\boldsymbol{v}_{\theta}$ is a function parameterized by $\theta$. Usually, $\boldsymbol{v}_{\theta}$ is the direct output of model~\cite{lipman2022flow,liu2022flow}. Velocity prediction can achieve strong performance for low-dimensional VAE latent~\cite{esser2024scaling,wan2025wan,flux}.

\paragraph{Representation autoencoder} RAE~\cite{zheng2025diffusion} uses pretrained representation encoders~(e.g., SigLIP2~\cite{tschannen2025siglip} and DINOv2~\cite{oquab2023dinov2}) to produce latents for diffusion models~\cite{rombach2022high,peebles2023scalable}. It presents promising performance for class-conditioned generation. Scale-RAE~\cite{tong2026scaling} scales further for text-to-image generation by using SigLIP-2~\cite{tschannen2025siglip}.

\paragraph{$\boldsymbol{x}_{0}$-prediction}Recently, JiT~\cite{li2025back} shows that standard velocity prediction struggles when data lies on a low-dimensional manifold within a high-dimensional space such as pixel space. Thus, it employs model to predict clean data $\boldsymbol{x}_{0}$ directly and velocity loss $\boldsymbol{v}$\textbf{-loss} by transformation: $\boldsymbol{v}_{\theta} = \frac{\boldsymbol{z}_{t}-\boldsymbol{x}_{\theta}}{t}$, where $\boldsymbol{x}_{\theta}$ is the model output, target is $\boldsymbol{v} = \frac{\boldsymbol{z}_{t}-\boldsymbol{x}_{0}}{t}$. The loss function becomes:
\begin{equation}
    \mathcal{L} = \mathbb{E}_{t, \boldsymbol{x}_{0}, \boldsymbol{\epsilon}} \| \boldsymbol{v}_\theta(\boldsymbol{z}_{t}, t) - \boldsymbol{v} \|^2 \\
    = \mathbb{E}_{t, \boldsymbol{x}_{0}, \boldsymbol{\epsilon}} \left\| \frac{\boldsymbol{x}_{0} - \boldsymbol{x}_{\theta}}{t}\right\|^2\text{.}
\end{equation}

\subsection{Strong-Reconstruction Representation Autoencoder}\label{method:effective-dimensionality}
\begin{table}[!t]
\centering
\small
\caption{\textbf{Effective dimensionality and reconstruction.} We report the effective dimensionality and reconstruction performance~(PSNR) on the ImageNet~\cite{deng2009imagenet} validation set. We evaluate two main groups of tokenizers/encoders. The first group utilizes checkpoints from RAE~\cite{zheng2025diffusion}, where all encoders are kept frozen and the decoders are trained on ImageNet~\cite{deng2009imagenet} (* indicates we trained the decoder for MAE-B from MAWS~\cite{singh2023effectiveness} on ImageNet using the same architecture). For the second group, we train two decoders separately: one with a frozen DINOv2-L~\cite{oquab2023dinov2} encoder and another with an unfrozen encoder following~\cite{chen2025aligning}. $R_{\tau}$ is defined in~\cref{eq:effective_dim}, which means minimum number of components required to account for at least $\tau \%$ of the variance.}

{\setlength{\tabcolsep}{4pt}
\begin{tabular}{l c c c c c c}
\toprule
\multirow{2}{*}{\textbf{Tokenizer/Encoder}}   & \multirow{2}{*}{\shortstack[c]{Training \\dataset}} &  \multicolumn{4}{c}{Dimensions}  & \multirow{2}{*}{\shortstack[c]{Reconstruction \\ (PSNR)}} \\
\cmidrule(lr){3-6}
& & Full & $R_{90}$ & $R_{95}$& $R_{99}$ & \\
\toprule
DINOv1-B  &  IN-1k &  768 & 502  & 595  &  701 & ---\\  
\cdashlinelr{1-7}
  DINOv2-B-RAE   &LVD-142M &  768 & 507 & 611 & 726 & 18.85 \\  
SigLIP2-B-RAE  & WebLI &  768  & 460   & 578 & 715 &  19.10 \\
  MAE-B-RAE &  IN-1k & 768 & 103  & 252  & 578 & 28.13 \\ 
  MAE-B-IG-3B*  & Instagram-3B & 768  & 197  & 348 &  595  & 27.67\\ 
\cdashlinelr{1-7}
  DINOv2-L & LVD-142M &  1024  & 672  & 811  &  965   & 17.34 \\  
  Finetuned DINOv2-L & --- & 1024  & 129  &302  &  726 & 29.12 \\ 
\bottomrule
\end{tabular}}
\label{tab:encoder_rank}
\end{table}

\begin{figure}
\centering

\includegraphics[width=\linewidth]{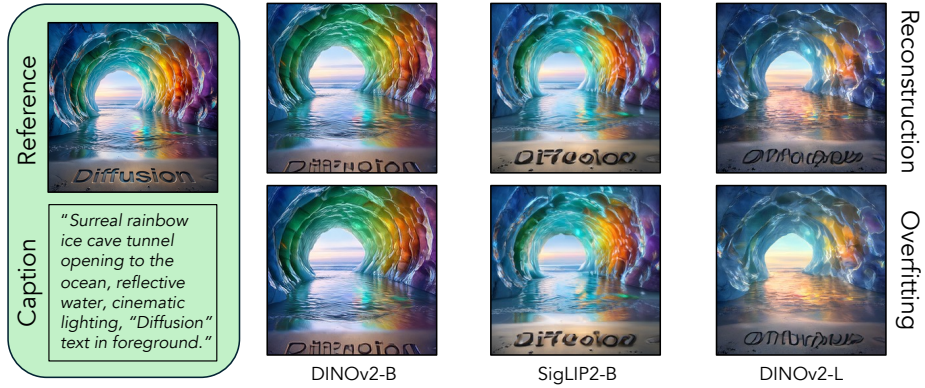}

\caption{\textbf{Reconstruction and overfitting for representation autoencoders.} We show reconstruction and overfitting results for DINOv2~\cite{oquab2023dinov2} and SigLIP-2~\cite{tschannen2025siglip}. These indicate that generation could be bottlenecked by the performance of reconstruction.}

\label{fig:recon_overfit}
\end{figure}
 
Representation Autoencoders (RAEs)~\cite{zheng2025diffusion,tong2026scaling} accelerate text-to-image diffusion training by adopting features from pretrained semantic encoders as their latent representation. Yet, these semantic latents can omit high-frequency information crucial for precise reconstruction. Our reconstruction and overfitting results in~\cref{fig:recon_overfit} suggest that reconstruction quality can limit generation fidelity in some cases. Consequently, we are motivated to explore generative modeling utilizing strong-reconstruction representation autoencoders, which also inherit semantic information. 

\paragraph{Effective dimensionality of representations}A straightforward approach to improving the reconstruction of the representation autoencoder is to finetune it using a reconstruction objective. However, as shown in~\cref{fig:teaser}, we find that it is challenging to perform standard flow matching $v$-prediction in the resulting representation space. Thus, we analyze the singular value spectrum of patch embeddings across different models. For each model, we extract L2-normalized per-patch features for 200k patches randomly sampled from ImageNet~\cite{deng2009imagenet}. Concretely, for each model, we subtract the mean feature vector across all sampled patches and apply singular value decomposition~(SVD) to the resulting centered patch embedding matrix $H\in\mathbb{R}^{N \times D}$ to obtain singular values $\sigma_{1},\sigma_{2},...,\sigma_{D}$ in non-increasing order, where $N$ is the number of patches and $D$ is the embedding dimension. The total energy is $E={\textstyle\sum}_{j=1}^{D}\sigma_{j}^2$, and fraction of energy explained by component $i$ is $f_{i} = \frac{\sigma_{i}^2}{E}$. The cumulative energy of the top $k$ components is: 
\begin{equation}
    C(k) = \sum\limits_{i=1}^{k}f_{i}= \frac{{\textstyle\sum}_{i=1}^{k}\sigma_{i}^{2}}{E}\text{.}
\end{equation}
Therefore, we define the effective dimensionality as: 
\begin{equation}\label{eq:effective_dim}
    R_{\tau} = \min \{k \mid C(k) \geq \frac{\tau}{100} \}.
\end{equation}This represents the minimum number of components required to account for at least $\tau\%$ of the variance in the feature embedding. In our analysis, we evaluate $\tau \in \{90, 95, 99\}$. We present the effective dimensionality and reconstruction performance on ImageNet~\cite{deng2009imagenet} validation set of two main groups of encoders/tokenizers in~\cref{tab:encoder_rank}.  

First, we reuse the decoder checkpoints from RAE~\cite{zheng2025diffusion}, where all encoders are kept strictly frozen and the decoders are trained on ImageNet~\cite{deng2009imagenet} for reconstruction. Additionally, we train a decoder for MAE-B from MAWS~\cite{singh2023effectiveness} on ImageNet~\cite{deng2009imagenet} using the same architecture. This MAE-B variant is pretrained on the much larger Instagram-3B dataset. Finally, following prior work~\cite{chen2025aligning}, we train two separate decoders for DINOv2-L~\cite{oquab2023dinov2} on ImageNet~\cite{deng2009imagenet} for image reconstruction: one where the DINOv2-L encoder remains strictly frozen, and another where the encoder is unfrozen and finetuned during training.

As shown in~\cref{tab:encoder_rank}, the effective dimensionality for the DINOv2~\cite{oquab2023dinov2} and SigLIP2~\cite{tschannen2025siglip} models remains high compared with the full feature dimension, which could partially explain their promising generation performance. DINOv1~\cite{caron2021emerging}, despite only being pretrained on ImageNet~\cite{deng2009imagenet}, also maintains a high effective dimensionality. In contrast, MAE~\cite{he2022masked}, whose pretraining objective is reconstruction, exhibits a much lower effective dimensionality, even when trained on a large-scale dataset~\cite{singh2023effectiveness}. Similarly, the finetuned DINOv2-L has a significantly lower effective dimensionality than the original frozen DINOv2-L~\cite{oquab2023dinov2}. These results indicate that when an encoder is trained for reconstruction, the ratio of effective dimensionality to the full feature dimension shrinks significantly.

\begin{figure}[!t]
\centering

\includegraphics[width=\linewidth]{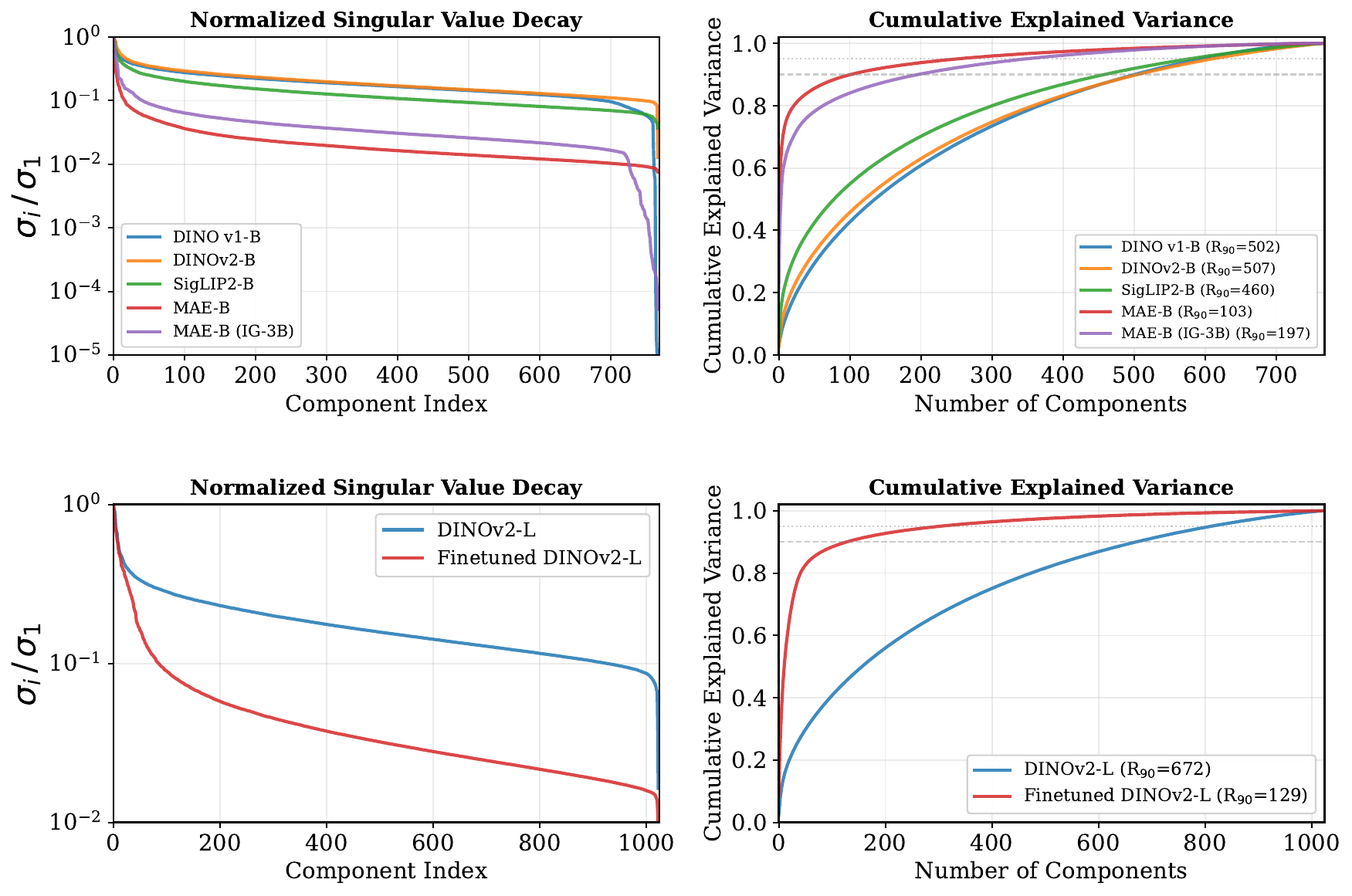}

\caption{\textbf{Effective dimensionality of vision encoder representations.} We analyze normalized singular value decay~(log-scale y-axis, normalized by $\sigma_{1}$) and cumulative explained variance for the visual encoders listed in~\cref{tab:encoder_rank} and observe that reconstruction-based objectives significantly reduce effective dimensionality. For instance, the normalized singular values of MAE~\cite{he2022masked} decay rapidly, even when pretrained on the massive Instagram-3B dataset. Similarly, when DINOv2-L~\cite{oquab2023dinov2} is finetuned via reconstruction, a much smaller number of principal components is required to capture the same amount of variance compared to the original model.}
\label{fig:svd}
\end{figure}

We further visualize this compression via normalized singular value decay and cumulative explained variance in~\cref{fig:svd}. As observed, the normalized singular values of MAE~\cite{he2022masked} decay rapidly, a bottleneck that persists even when pretrained on the massive Instagram-3B dataset. Similarly, when DINOv2-L is finetuned via reconstruction, a much smaller number of principal components is required to capture the same amount of variance compared to the original model. A potential reason for this is that to perform reconstruction, the encoder must capture a significant amount of high-frequency information from the original image. Since natural images are suggested to reside on a low-dimensional manifold~\cite{li2025back,Chapelle2006SemiSupervised,carlsson2009topology}, the features are forced to mimic this lower-dimensional distribution.

Based on~\cref{tab:encoder_rank} and~\cref{fig:svd}, our hypothesis is that high-dimensional encoders that are trained or finetuned for image reconstruction objective will concentrate near a low-dimensional subspace. Assume the observed high-dimensional feature $\boldsymbol{x}_{0} \in \mathbb{R}^{h}$, extracted by a visual encoder, lies within a lower-dimensional subspace. Specifically, $\boldsymbol{x}_{0}$ is generated from a true latent variable $\boldsymbol{c}_{0} \in \mathbb{R}^{l}$ via the linear mapping $\boldsymbol{x}_{0} = Q\boldsymbol{c}_{0}$, where the columns of $Q \in \mathbb{R}^{h \times l}$ form an orthonormal basis for the subspace (i.e., $Q^T Q = I_l$). Input $\boldsymbol{z}_{t}^{h} = (1-t)\boldsymbol{x}_{0} + t\boldsymbol{\epsilon}_{h}$ could be decomposed into $\boldsymbol{z}_{t}^{h} = Q ((1-t)\boldsymbol{c}_{0} + t\boldsymbol{\epsilon}_{l} ) + t\boldsymbol{\epsilon}_\perp$, where $\boldsymbol{\epsilon}_\perp = \boldsymbol{\epsilon}_{h} - Q\boldsymbol{\epsilon}_{l} = (I-QQ^{T})\boldsymbol{\epsilon}_{h}$. $\boldsymbol{z}_{t}^{l} = (1-t)\boldsymbol{c}_{0} + t\boldsymbol{\epsilon}_{l}$ can be viewed as manifold component. $\boldsymbol{\epsilon}_{\perp}$ is the normal component and we have $(I-QQ^{T})\boldsymbol{z}_{t}^{h} = t\boldsymbol{\epsilon}_{\perp}$. The model that predicts velocity in high-dimensional space is defined as: $\boldsymbol{v}_{\theta}^{h}(\boldsymbol{z}_{t}^{h}, t) = \mathbb{E}[\boldsymbol{\epsilon_{h}} - \boldsymbol{x}_{0} | \boldsymbol{z}_{t}^{h}] = \mathbb{E} [ Q (\boldsymbol{\epsilon}_{l} - \boldsymbol{c}_{0}) + \boldsymbol{\epsilon}_{\perp} | \boldsymbol{z}_{t}^{h}]$. Therefore, we can obtain
 \begin{equation}
\begin{aligned}
    \boldsymbol{v}_{\theta}^{h}(\boldsymbol{z}_t^{h}, t) &= \mathbb{E} [ Q (\boldsymbol{\epsilon}_{l} - \boldsymbol{c}_{0}) | \boldsymbol{z}_{t}^{h}] + \mathbb{E} [ \boldsymbol{\epsilon}_{\perp} | \boldsymbol{z}_{t}^{h}] \\
    &= Q\boldsymbol{v}_{\theta}^{l}(Q^\top \boldsymbol{z}_t^{h}, t) + \frac{1}{t}(I - QQ^\top)\boldsymbol{z}_{t}^{h}\text{.}
\end{aligned}
\label{eq:velocity}
\end{equation}
When the dimension of $\boldsymbol{x}_{0}$ is much higher than the dimension of $\boldsymbol{c}_{0}$~($h \gg l$), the model needs to allocate substantial capacity for $\frac{1}{t}(I - QQ^\top)\boldsymbol{z}_{t}^{h}$. As discussed in~\cite{zhang2025both}, one solution is to train an adapter to reduce dimensionality. Instead, we focus on predicting high-dimensional features in this paper, as retaining dimensionality might be able to retain both rich semantic information and reconstruction capability. The second term in~\cref{eq:velocity} is from $\boldsymbol{\epsilon}_{\perp}$, which is introduced by $\boldsymbol{\epsilon}_{h}$ in the vanilla velocity prediction training target. To avoid this, we employ the model to predict $\boldsymbol{x}_{0}$ directly:
\begin{equation}
    \boldsymbol{x}_{\theta}^{h}(\boldsymbol{z}_{t}^{h}, t)  = \mathbb{E}[\boldsymbol{x}_{0} | \boldsymbol{z}_{t}^{h}] = \mathbb{E}[Q\boldsymbol{c}_{0}|Q\boldsymbol{z}_{t}^{l} + t\boldsymbol{\epsilon}_{\perp}] = \mathbb{E}[Q\boldsymbol{c}_{0}|Q\boldsymbol{z}_{t}^{l},  t\boldsymbol{\epsilon}_{\perp}].
\label{eq:x0-pred}
\end{equation}
Since $Q\boldsymbol{c}_{0}$ is independent of $t\epsilon_{\perp}$, \cref{eq:x0-pred} could be simplified to:
\begin{equation}
    \boldsymbol{x}_{\theta}^{h}(\boldsymbol{z}_{t}^{h}, t) = \mathbb{E}[Q\boldsymbol{c}_{0} | Q\boldsymbol{z}_{t}^{l}] =Q\mathbb{E}[\boldsymbol{c}_{0} | Q\boldsymbol{z}_{t}^{l}]=Q\boldsymbol{x}_{\theta}^{l}(\boldsymbol{z}_{t}^{l}, t).
\label{eq:x0-pred-simplified}
\end{equation}
\cref{eq:x0-pred-simplified} shows that directly modeling $\boldsymbol{x}_{0} \in \mathbb{R}^{h}$ in high-dimensional space could predict $\boldsymbol{c}_{0} \in \mathbb{R}^{l}$ more efficiently, especially when $h \gg l$.

\paragraph{$\boldsymbol{x}_{0}$-prediction for representation latent} Based on our analysis above, we prefer to do $\boldsymbol{x}_{0}$-prediction for strong-reconstruction representation autoencoders. Specifically, given text-image pair~$(\boldsymbol{I}_{i}, \boldsymbol{T}_{i})$, extracted text embeddings act as the condition $\boldsymbol{y}_{i}$ for diffusion transformer $\boldsymbol{x}_{\theta}$. Feature $\boldsymbol{h}_{i}$ produced by representation autoencoder $g$: $\boldsymbol{h}_{i} = g(\boldsymbol{I}_{i})$ is used as latent for diffusion. Given input $\boldsymbol{h}_{i,t} = (1-t)\boldsymbol{h}_{i,0} + t\boldsymbol{\epsilon}$, DiT is trained to denoise noisy image representations. The loss function is defined as follows:
\begin{equation}
    \mathcal{L}_{\text{Diff}} 
    = \mathbb{E}_{t, \boldsymbol{h}_{i,0}, \boldsymbol{\epsilon}} \left\| \frac{\boldsymbol{h}_{i, 0} - \boldsymbol{x}_{\theta}(\boldsymbol{h}_{i,t}, t, \boldsymbol{y}_{i})}{t}\right\|^2\text{,}
\end{equation}
where $\boldsymbol{h}_{i,0}$ means the clean feature $\boldsymbol{h}_{i,0} \coloneqq \boldsymbol{h}_{i}$. In practice, we clamp $t$ to a minimum of 0.05, following~\cite{li2025back}, to prevent numerical instability.

\begin{figure}[!t]
\centering

\includegraphics[width=\linewidth]{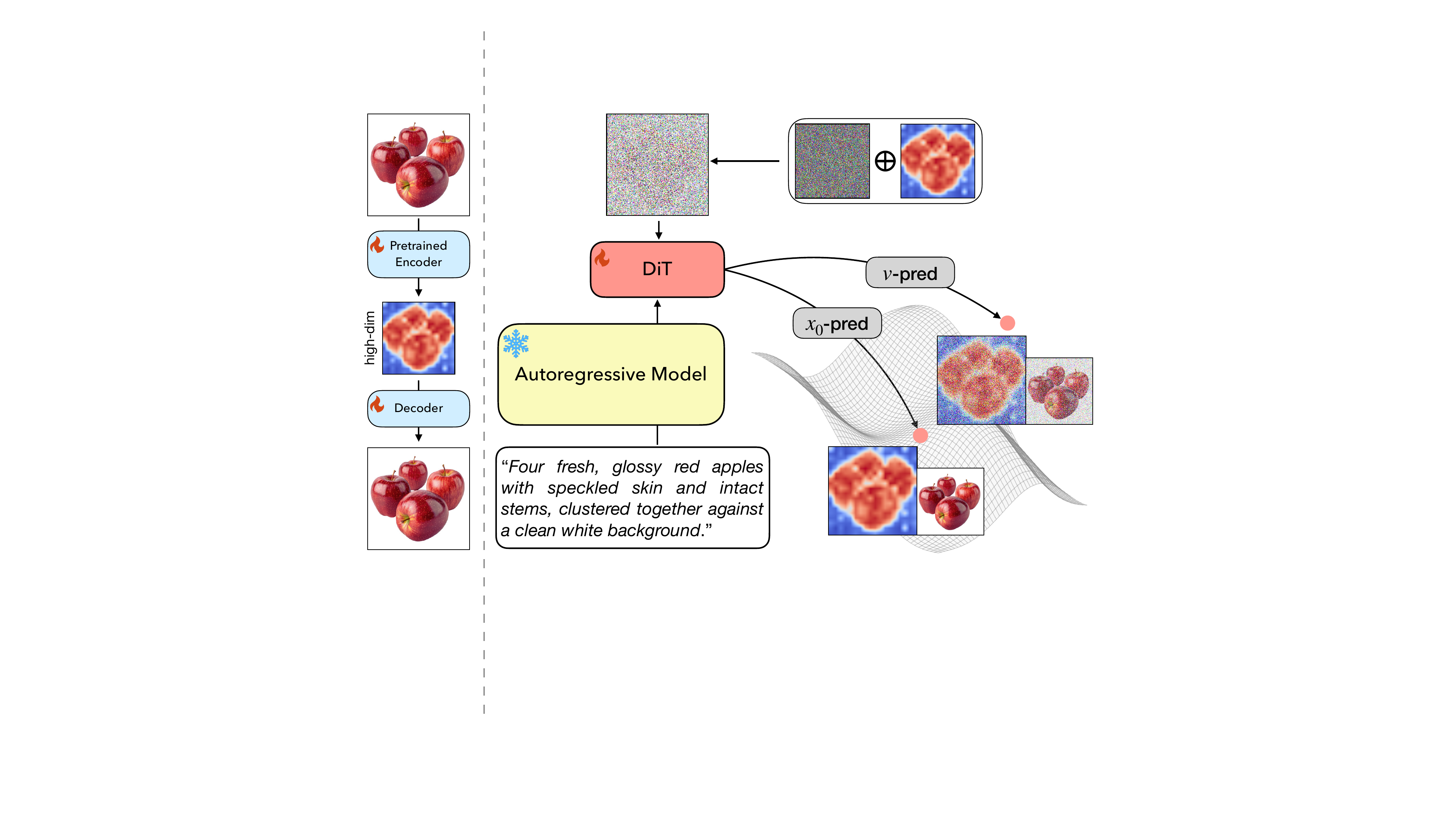}
    \begin{flushleft}
        \vspace{-3mm}
  \hspace{-1mm}   (a)  Tokenizer \hspace{20mm} (b) Text-to-image generation framework
         \vspace{-5mm}
    \end{flushleft} 
\caption{\textbf{Method.} (a) We unfreeze pretrained visual encoder when training the decoder for reconstruction. Then we use high-dimensional representation from finetuned encoder as latent. (b) We use autoregressive model and query tokens to extract text embedding as conditioning for DiT. Due to the low-dimensional manifold property, we employ $\boldsymbol{x}_{0}$ prediction to train text-to-image DiT more efficiently.}

\label{fig:overall}
\end{figure}

\paragraph{Text-to-image generation architecture} 
As shown in~\cref{fig:overall}, we mainly focus on using representation autoencoder with strong reconstruction performance for text-to-image generation. The high-dimensional visual features produced from it are used as latents for diffusion. For the generation framework, we adopt an architecture similar to MetaQuery~\cite{pan2025transfer} and BLIP3-o~\cite{chen2025blip3}, where we use a frozen pretrained vision language model~(VLM) for the autoregressive model. Learnable query tokens are employed to extract text information as conditioning for randomly initialized diffusion transformer~(DiT).

\section{Experiments}
\subsection{Implementation Details}
We use Qwen3-VL-2B-Instruct~\cite{bai2025qwen3} as our~(frozen) autoregressive model and Lumina-Next~\cite{zhuo2024lumina} with 1B size as our DiT. The DiT consists of 24 layers and 24 attention heads per layer, with a hidden dimension of 1536. The number of query tokens is set to be 64. We use AdamW~\cite{loshchilov2017decoupled} with the learning rate of $10^{-4}$. Global batch size is 1024. We use a 34M subset of the public pretraining dataset used in BLIP-3o~\cite{chen2025blip3}. All text-to-image generation models are trained for 90k steps. The whole size of the public BLIP-3o~\cite{chen2025blip3} pretraining dataset is 39.3M. It combines mostly webdata like CC12M~\cite{changpinyo2021conceptual}, SA-1B~\cite{kirillov2023segment}, and JourneyDB~\cite{sun2023journeydb}, and recaptions many images. VLM is frozen during training. DiT and learnable query tokens are trained from scratch. Following prior work~\cite{zheng2025diffusion,tong2026scaling}, we adopt the uniform time schedule with a shift for DiT training. Concretely, given base $t_{n}$, the shifted $t_{m}$ is defined as follows:
\begin{equation}
    t_{m} = \frac{\alpha t_{n}
    }{1 + (\alpha -1)t_{n}}\text{,}
\end{equation}
where $\alpha = \sqrt{\frac{m}{n}}$, $m = \text{number of image tokens} \times \text{token dimension}$. We use $n=4096$ following RAE~\cite{zheng2025diffusion,tong2026scaling}. We use two tokenizers: 1) frozen MAE pretrained on ImageNet~\cite{deng2009imagenet} paired with a ViT-XL~\cite{dosovitskiy2020image} decoder from RAE~\cite{zheng2025diffusion}, which is also trained on IN-1k~\cite{deng2009imagenet}, and 2) we unfreeze DINOv2-L~\cite{oquab2023dinov2} during reconstruction training on IN-1k~\cite{deng2009imagenet}, which is paired with a decoder with the similar architecture as VAE used in stable diffusion~\cite{rombach2022high} following~\cite{chen2025aligning}.

When we finetune DINOv2-L~\cite{oquab2023dinov2} encoder, we also leverage a semantic preservation loss, in addition to the reconstruction loss, to prevent collapse~\cite{chen2025aligning}. Specifically, for each image $I_{i}$, feature $\boldsymbol{h}_{i}$ is extracted by original DINOv2-L~\cite{oquab2023dinov2} $g$ for semantic preservation. The final loss to finetune encoder DINOv2-L $g\prime$ is defined as follows:
\begin{equation}
    \mathcal{L}_{FT} = \|g(I) - g\prime(I)\|^2  + \mathcal{L}_{\text{recon}} ,
\end{equation}
where $\mathcal{L}_{\text{recon}}$ consists of pixel-level $L1$, perceptual, and adversarial losses. Unless otherwise specified, all experiments are conducted at a resolution of $256 \times 256$. The model training is using either 64 NVIDIA A100 or 32 H200 GPUs.

\subsection{Evaluation}
We evaluate FID~\cite{heusel2017gans} on COCO-30k~\cite{lin2014microsoft} for image quality. Following prior work~\cite{chen2025blip3,tong2026scaling}, we also use two widely adopted metrics: the GenEval score~\cite{ghosh2023geneval} and the DPG-Bench score~\cite{hu2024ella} for evaluation of text-image alignment. We employ a 100-step Euler sampler for generation. 

\subsection{Text-to-image Generation Comparison} 
\paragraph{$\boldsymbol{x}_{0}$-prediction better than $\boldsymbol{v}$-prediction for strong-reconstruction representation autoencoders}

\begin{table}[!t]
\centering
\small
\setlength{\tabcolsep}{2pt}   %
\caption{\textbf{Evaluation of finetuned DINOv2-L.} We compare the text-to-image generation performance of the original DINOv2-L~\cite{oquab2023dinov2} against our finetuned variants on three benchmarks: GenEval~\cite{ghosh2023geneval}, DPG-Bench~\cite{hu2024ella}, and COCO-30k FID~\cite{lin2014microsoft}. Following RAE~\cite{zheng2025diffusion}, we use standard $\boldsymbol{v}$-prediction for DINOv2-L. For the fine-tuned models, we evaluate  $\boldsymbol{v}$ and $\boldsymbol{x}_{0}$ prediction targets. The FT-DINOv2-L-low-dim variant utilizes a trained adapter to down-project high-dimensional features, performing standard flow matching in this lower-dimensional space. Reconstruction performance is measured by PSNR. Best and second-best are highlighted in \textbf{bold} and \textcolor{blue}{blue}, respectively.}
\begin{tabular}{l c c c c c c}
\toprule
\textbf{Tokenizer}     & Dim. & \begin{tabular}{@{}c@{}}Pred. \\ type\end{tabular} &  GenEval$\uparrow$ & DPG-Bench$\uparrow$   & coco-30k FID$\downarrow$ & Recon.$\uparrow$ \\
\toprule
FT-DINOv2-L-low-dim  & 32   &  $\boldsymbol{v}$ & \textbf{40.99} & \textbf{73.04} &  \textcolor{blue}{17.64} & \textcolor{blue}{25.83} \\
\cdashlinelr{1-7}
DINOv2-L  & 1024  &  $\boldsymbol{v}$ & 37.63 &  70.21  & 18.34 & 17.34 \\
\cdashlinelr{1-7}
 FT-DINOv2-L   &   1024 & $\boldsymbol{v}$  & 30.96 &  67.47 & 29.97 & \multirow{2}{*}{\textbf{29.12}}  \\
  FT-DINOv2-L & 1024  & $\boldsymbol{x}_{0}$  & \textcolor{blue}{39.48} &  \textcolor{blue}{71.83} & \textbf{16.80} &  \\
\bottomrule
\end{tabular}
\label{tab:dino_comparison}

\end{table}

To evaluate the finetuned DINOv2-L tokenizer, we compare four configurations: (1) the original DINOv2-L~\cite{oquab2023dinov2} tokenizer using $\boldsymbol{v}$-prediction, following RAE~\cite{zheng2025diffusion}; (2) the finetuned DINOv2-L tokenizer using standard $\boldsymbol{v}$-prediction; (3) the finetuned DINOv2-L tokenizer using $\boldsymbol{x}_{0}$-prediction; and (4) a low-dimensional variant. For this fourth setup, we train an adapter to down-project the high-dimensional (1024-dim) features into a 32-dimensional space, perform standard flow matching ($\boldsymbol{v}$-prediction) in this compressed space, and train the decoder specifically on these 32-dimensional features rather than the full 1024 dimensions following~\cite{chen2025aligning}. We present result in~\cref{tab:dino_comparison}. When we finetune DINOv2-L~\cite{oquab2023dinov2} by reconstruction objective and perform standard $\boldsymbol{v}$-prediction, we observe performance deterioration on three benchmarks: GenEval~\cite{ghosh2023geneval}, DPG-Bench~\cite{hu2024ella}, and COCO-30k FID~\cite{lin2014microsoft} as suggested by our analysis in~\cref{method:effective-dimensionality}. Even though unfreezing the encoder should let the encoder capture more information about the image, the diffusion model seems to struggle to denoise the finetuned features correctly. This suggests that finetuning the encoder by reconstruction worsens the diffusibility of representation space when using standard flow matching $\boldsymbol{v}$-prediction. 

\begin{figure}[!t]
\centering

\includegraphics[width=\linewidth]{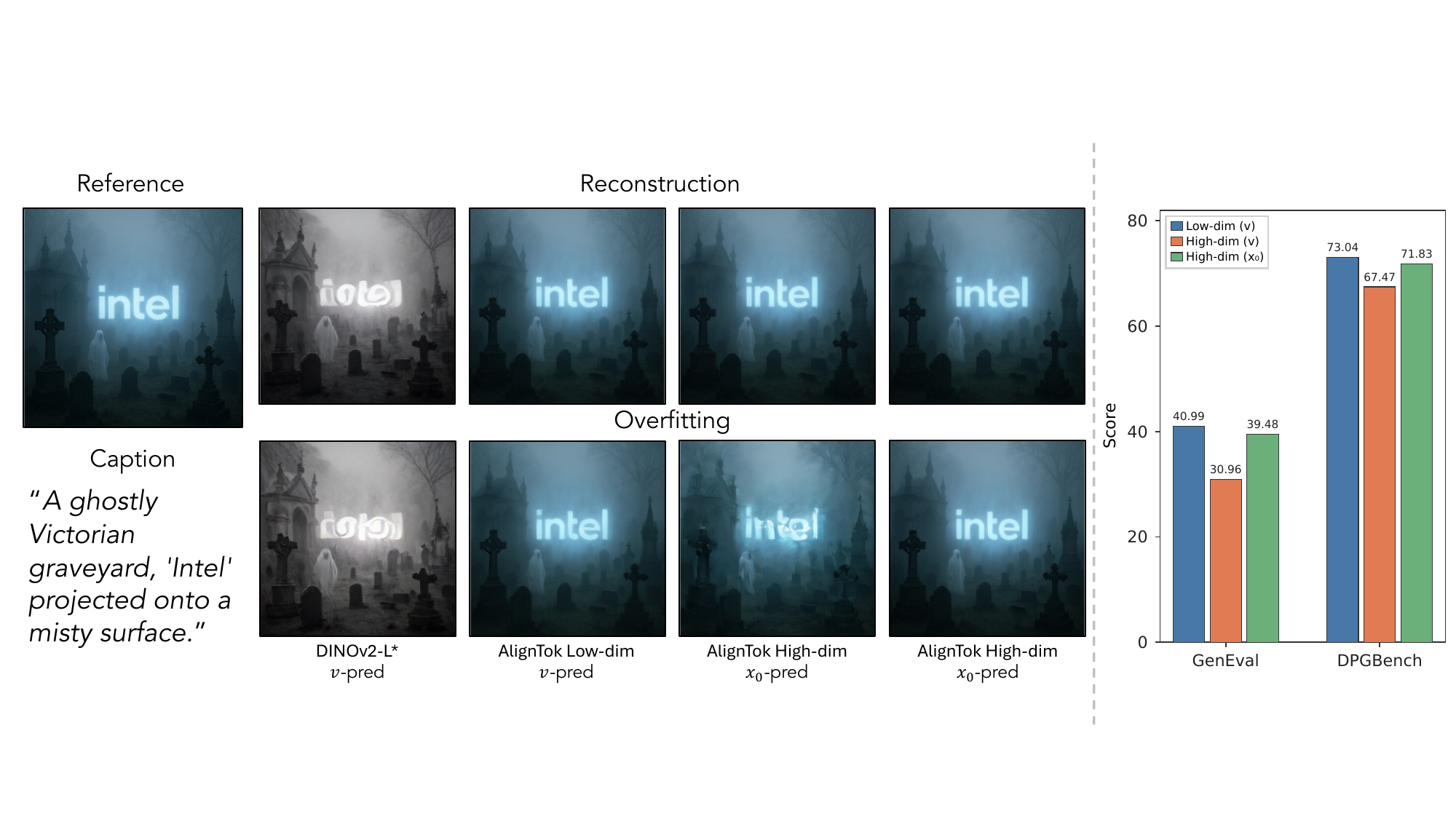}

\caption{\textbf{Qualitative comparison of $\boldsymbol{x}_{0}$ and $\boldsymbol{v}$ predictions for finetuned DINOv2-L.} We present text-to-image generation examples comparing the $\boldsymbol{x}_{0}$ and $\boldsymbol{v}$ prediction targets using the finetuned DINOv2-L tokenizer. The results demonstrate that $\boldsymbol{x}_{0}$ prediction yields better visual quality.}
\label{fig:aligntok-compare}
\end{figure}

However, switching to $\boldsymbol{x}_{0}$-prediction boosts performance by a relatively large margin and surpasses original DINOv2~\cite{oquab2023dinov2} with $\boldsymbol{v}$-prediction. This solidifies our analysis in~\cref{method:effective-dimensionality}, when effective dimensionality of visual representation is low~(e.g., finetuned DINOv2 in~\cref{tab:encoder_rank}), using $\boldsymbol{x}_{0}$-prediction leads to faster convergence for generative models. We present some qualitative results in~\cref{fig:aligntok-compare} to compare $\boldsymbol{x}_{0}$-prediction and $\boldsymbol{v}$-prediction for finetuned DINOv2 tokenizer. \cref{fig:aligntok-compare} shows that $\boldsymbol{x}_{0}$-prediction leads to better overall visual quality, such as object structure and text alignment. For example, given text prompt ``{\tt a photo of a pink car}'', sampled image by $\boldsymbol{v}$-prediction struggles to capture the global geometry of the object, resulting in amorphous, blob-like structures. In contrast, $\boldsymbol{x}_0$-prediction maintains strong structural integrity and generates geometrically coherent objects.

\begin{figure}[t]
\centering

\includegraphics[width=\linewidth]{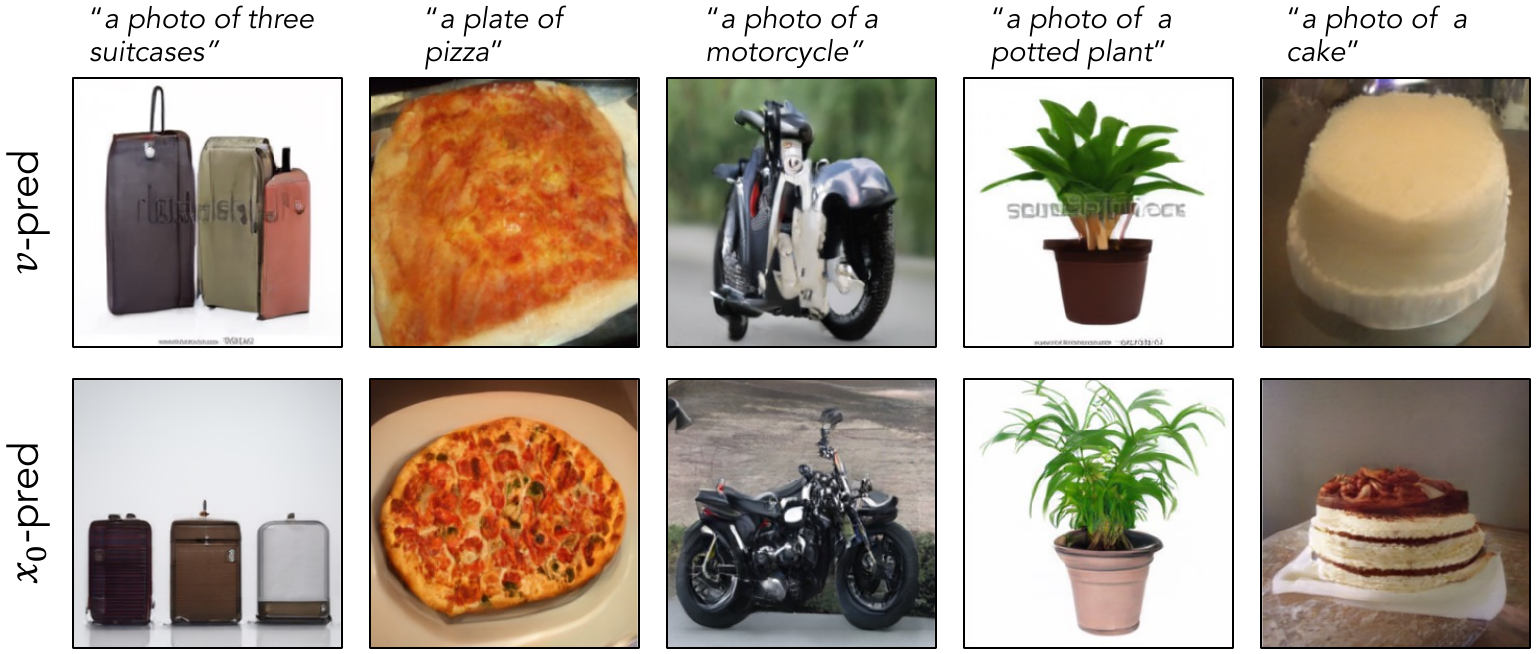}

\caption{\textbf{Qualitative comparison of $\boldsymbol{x}_{0}$ and $\boldsymbol{v}$ predictions for MAE.} We present generated examples comparing the $\boldsymbol{x}_{0}$ and $\boldsymbol{v}$ prediction targets using the MAE-RAE tokenizer~\cite{zheng2025diffusion}. The results demonstrate that $\boldsymbol{x}_{0}$ prediction yields better visual quality.}

\label{fig:mae-compare}
\end{figure}

Additionally, we present results for tokenizer MAE-RAE~\cite{zheng2025diffusion} in~\cref{tab:mae_comparison}, where encoder MAE is pretrained on IN-1k~\cite{deng2009imagenet} and kept frozen during training of decoder. The pretraining objective for encoder MAE is reconstruction, it has low effective dimensionality compared with its own full feature dimension as suggested in~\cref{tab:encoder_rank}. As shown in~\cref{tab:mae_comparison}, $\boldsymbol{x}_{0}$-prediction improves text-to-image generation performance on three benchmarks in comparison with $\boldsymbol{v}$-prediction. The qualitative results in~\cref{fig:mae-compare} show that sampled images by $\boldsymbol{x}_{0}$-prediction have better object structure and richer details. These results suggest that when high-dimensional visual feature/latent resides on a low-dimensional manifold, using $\boldsymbol{x}_{0}$-prediction can make diffusion models on this high-dimensional representation space converge faster and yield better sampled results. 

\begin{figure}[t]
\centering

\includegraphics[width=\linewidth]{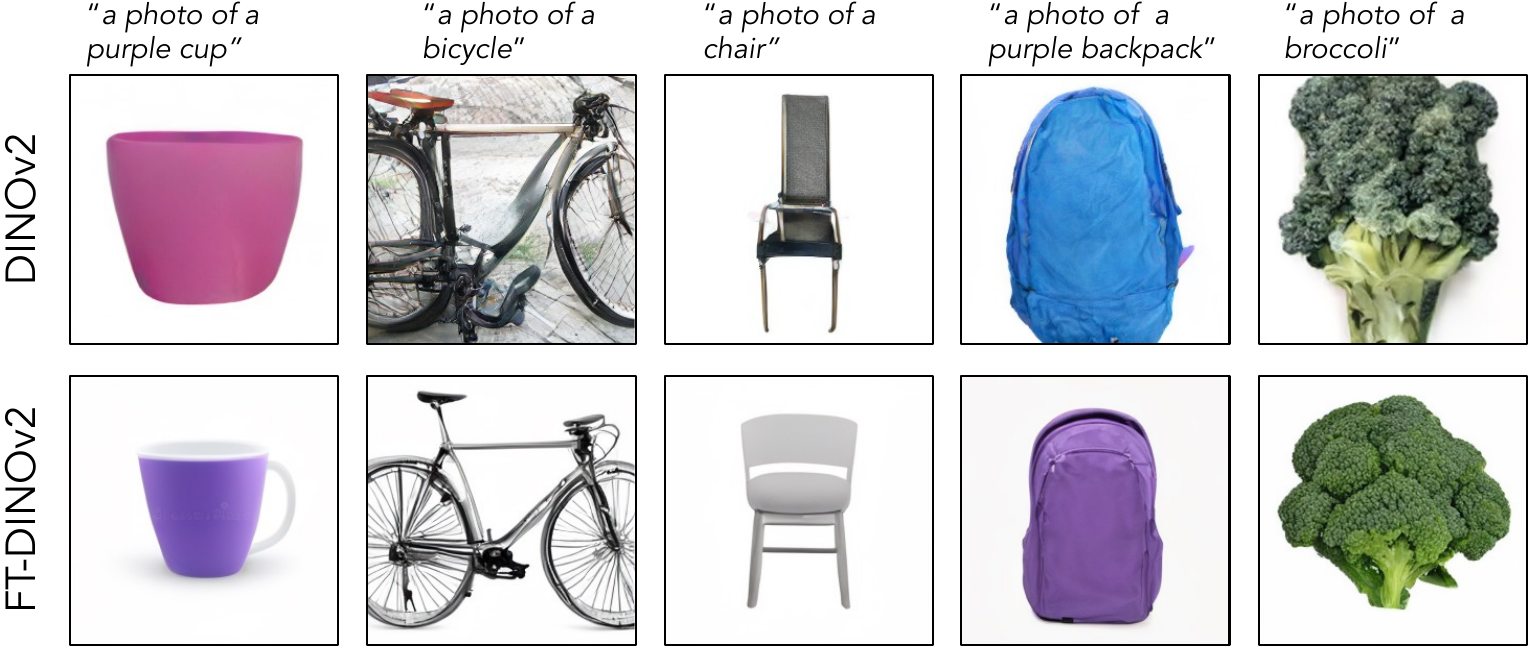}

\caption{\textbf{Qualitative comparison of DINOv2-L and finetuned DINOv2-L.} We present text-to-image generation examples comparing the tokenizers of the original DINOv2-L~\cite{oquab2023dinov2} with $\boldsymbol{v}$-prediction (top) and our finetuned DINOv2-L variant with $\boldsymbol{x}_{0}$-prediction (bottom). The results indicate that the fine-tuned encoder leads to better structural integrity and color alignment.}

\label{fig:dinov2-finetuned-compare}
\end{figure}

\paragraph{Reconstruction still matters for generation}
The semantic space formed by pretrained visual encoders like DINOv2~\cite{oquab2023dinov2} can lift a lot for generation. Semantics help models converge faster~\cite{zheng2025diffusion,tong2026scaling}. To investigate whether reconstruction can help generation, we compare tokenizer of original DINOv2-L~\cite{oquab2023dinov2} with tokenizer of finetuned DINOv2-L. For original DINOv2~\cite{oquab2023dinov2}, we use $\boldsymbol{v}$-prediction. $\boldsymbol{x}_{0}$-prediction is employed for finetuned DINOv2-L since previous section suggests that $\boldsymbol{v}$-prediction struggles to diffuse efficiently for finetuned DINOv2-L. Quantitative and qualitative results are presented in~\cref{tab:dino_comparison} and~\cref{fig:dinov2-finetuned-compare}, respectively. \cref{tab:dino_comparison} shows that finetuned DINOv2-L achieves better results on three benchmarks: GenEval~\cite{ghosh2023geneval}, DPG-Bench~\cite{hu2024ella}, and COCO-30k FID~\cite{lin2014microsoft}. As presented in~\cref{fig:dinov2-finetuned-compare}, diffusion model based on original DINOv2-L~\cite{oquab2023dinov2} fails to capture correct color information~(e.g., {\tt "a photo of a purple backpack"}) or the object integrity when texture is complicated~(e.g., {\tt "a photo of a bicycle"}). Though Scale-RAE~\cite{tong2026scaling} presents that scaling the size of training dataset can alleviate this problem, our experiments show the certain limitation of using features from pretrained semantic encoders without reconstruction tuning. We will explore more regarding building visual encoder for generative and unified models~(understanding and generation) in the future.

\begin{table}[t]
\centering
\small
\caption{\textbf{Evaluation of MAE-RAE.} We compare the text-to-image generation performance of $\boldsymbol{x}_{0}$ and $\boldsymbol{v}$ predictions for MAE-RAE~\cite{zheng2025diffusion} across three benchmarks: GenEval~\cite{ghosh2023geneval}, DPG-Bench~\cite{hu2024ella}, and COCO-30k FID~\cite{lin2014microsoft}. Best are highlighted in \textbf{bold}.}
\begin{tabular}{l c c c c c }
\toprule

\textbf{Tokenizer}     & Dimension & \begin{tabular}{@{}c@{}}Prediction \\ type\end{tabular} &  GenEval$\uparrow$ & DPG-Bench$\uparrow$   & COCO-30k FID$\downarrow$ \\
\toprule  
MAE-B-RAE  & 768   &   $\boldsymbol{v}$ & 36.17 & 67.54 &  22.20 \\  
MAE-B-RAE  & 768   &   $\boldsymbol{x}_{0}$ & \textbf{40.89} & \textbf{73.90} & \textbf{17.24} \\
\bottomrule
\end{tabular}
\label{tab:mae_comparison}
\vspace{1mm}
\end{table}

\paragraph{High-dimensional diffusion is inherently harder than the low-dim} Following~\cite{chen2025aligning}, we train another variant for finetuned DINOv2-L. Concretely, we set a two-layer MLP after encoder to down project high-dimensional feature~(1024-dim) to be low-dimensional~(32-dim), the subsequent decoder is also based on 32-dim features to do pixel decoding. We use $\boldsymbol{v}$-prediction following standard practice for this low-dimensional latent space. We present results in~\cref{tab:dino_comparison}. While the low-dimensional bottleneck (FT-DINOv2-L-low-dim) achieves stronger text-alignment metrics (GenEval~\cite{ghosh2023geneval} and DPG-Bench~\cite{hu2024ella}), this compression comes at the cost of image quality, yielding a worse FID score~(17.64). In contrast, our unbottlenecked high-dimensional approach with $\boldsymbol{x}_0$-prediction avoids this dimensionality reduction, achieving better visual fidelity~(FID 16.80) while maintaining competitive semantic alignment. This demonstrates that $\boldsymbol{x}_0$-prediction successfully mitigates the optimization difficulty of high-dimensional spaces, preserving the rich, fine-grained details of strong-reconstruction encoders.

{\setlength{\tabcolsep}{3pt}
\begin{table}[!h]
\centering
\small
\caption{\textbf{Scaling to a larger dataset.}
We train our method on a larger-scale dataset and perform supervised fine-tuning (SFT)
at resolutions of 256 and 512. The asterisk (*) denotes our own SFT implementation.}
\resizebox{\linewidth}{!}{%
\begin{tabular}{l c c c c c c}
\toprule
\textbf{Method} & DiT size & Training dataset & Latent dim. & Res. & GenEval$\uparrow$ & DPG-Bench$\uparrow$ \\
\midrule
Ours & 1B & 90M & 1024 & 256 & 43.84 &  76.15 \\
Ours & 1B & 90M + SFT 60k & 1024 & 256 & 78.62 &  80.13 \\
Ours~(512) & 1B & 90M~(256) + SFT 60k~(512) & 1024 & 512 & \textbf{79.69}  &  \textbf{81.15} \\
\cdashlinelr{1-7}
Scale-RAE*~\cite{tong2026scaling} & 2.4B & 64M + SFT 60k & 1152 & 224 & 77.43 & 78.47  \\
\bottomrule
\end{tabular}%
}
\label{tab:method_comparison}
\end{table}}

\paragraph{Scaling to a Larger Dataset}
We further train a diffusion model with $\boldsymbol{x}_{0}$-prediction using the
finetuned DINOv2 tokenizer on a larger internal 90M dataset, while keeping the
remaining hyperparameters unchanged. Following Scale-RAE~\cite{tong2026scaling},
we then finetune the pretrained model on BLIP3o-60k~\cite{chen2025blip3} at
256 resolution, and additionally finetune a high-resolution variant at 512 resolution.
We include Scale-RAE~\cite{tong2026scaling} as a reference. Results are shown
in~\cref{tab:method_comparison}. These results suggest that our method has potential scalability and could achieve competitive performance for text-to-image generation.

\section{Conclusion}
In this paper, we investigate text-to-image generation in strong-reconstruction, high-dimensional representation spaces. We show that after tuning the encoder by reconstruction objective, the standard velocity prediction becomes difficult to optimize in the resulting high-dimensional representation space. Empirically, reconstruction-tuned representation autoencoders exhibit effective-dimensionality collapse. We provide an analysis explaining this failure mode and propose a simple remedy: training the diffusion model to predict the clean representation $\boldsymbol{x}_{0}$ directly. Across two strong-reconstruction tokenizers, $\boldsymbol{x}_{0}$-prediction consistently improves text-to-image generation quality.

\paragraph{Acknowledgements} We thank Sicheng Mo, Xichen Pan, Eli Shechtman, Krishna Kumar Singh, Phillip Isola, and Nicolas Dufour for their valuable discussions and help. This work was supported in part by NSF CAREER $\#$2339071.

\clearpage
\bibliographystyle{splncs04}
\bibliography{main}
\clearpage
\appendix
\renewcommand{\thesection}{A.\arabic{section}}
\setcounter{section}{0}

\section{Implementation Details}
\paragraph{Tokenizer training}
Following~\cite{chen2025aligning}, we use three stages to finetune DINOv2-L~\cite{oquab2023dinov2} and train the decoder by the reconstruction objective. Specifically, we use four losses: 1) semantic preservation; 2) L1 loss; 3) perceptual loss; and 4) adversarial loss~\cite{rombach2022high,kouzelis2025eq,wen2025principal,wu2025towards,goodfellow2020generative,dosovitskiy2020image,skorokhodov2025improving,hansen2025learnings,silvestri2025covae,van2017neural,esser2021taming,johnson2016perceptual,gatys2015neural,zhang2018unreasonable}. For each image $I_{i}$, feature $\boldsymbol{h}_{i}$ is extracted by original encoder $g$ for semantic preservation $\boldsymbol{h}_{i} = g(I_{i})$. The final loss for the finetuned encoder $g\prime$ and decoder $D$ is defined as follows:
\begin{equation}
    \mathcal{L}_{FT} = \omega_{sp}\|\boldsymbol{h}_{i}^{\prime} - \boldsymbol{h}_{i}\|^2  + \mathcal{L}_{\text{L1}}(I_{i}^{\prime}, I_{i}) + \omega_{p}\mathcal{L}_{\text{perceptual}}(I_{i}^{\prime}, I_{i}) + \omega_{g}\mathcal{L}_{\text{GAN}}(I_{i}^{\prime}, I_{i})  ,
\end{equation}
where $\boldsymbol{h}_{i}^{\prime} = g\prime(I_{i})$ and $I_{i}^{\prime} = D(\boldsymbol{h}_{i}^{\prime})$. We set $\omega_{sp} = 1, \omega_{p}=1, \text{and } \omega_{g}=0.5$ following~\cite{chen2025aligning}.

\begin{table}[h]
    \caption{{\bf Optimization hyperparameters for the 1B Diffusion Transformer}.}
    \footnotesize
    \centering
    \begin{tabular}{lc}
    \toprule
    Hyperparameter & Value \\
    \midrule
    Optimizer & AdamW \\
    Max learning rate & $1 \times 10^{-4}$ \\
    Min learning rate & $1 \times 10^{-5}$ \\
    Learning rate schedule & Cosine \\
    Warmup ratio & 0.003 \\
    $\beta_1, \beta_2$ & 0.9, 0.999 \\
    Global batch size & 1024 \\
    Gradient norm clip & 1.0 \\
    \bottomrule
    \end{tabular}
    \label{tab:optim-hyperparams}
\end{table}

\paragraph{DiT Architecture} 
The DiT backbone consists of 24 layers with 24 attention heads each, and a hidden dimension of 1536. We utilize the SwiGLU~\cite{shazeer2020glu} activation function. The Classifier-Free Guidance (CFG) dropout probability is set to $0.1$. The input latent shapes for the DiT are $1024 \times 16 \times 16$ when using the finetuned DINOv2 encoder and $768 \times 16 \times 16$ for the MAE encoder~\cite{he2022masked}.

\paragraph{Optimization} 
Detailed optimization hyperparameters are provided in~\cref{tab:optim-hyperparams}.

\section{Analysis of Effective Dimensionality on More Datasets}
 Effective dimensionality for tokenizers/encoders is evaluated on three additional larger datasets~(ImageNet-21k~\cite{ridnik2021imagenetk}, BLIP3o~\cite{chen2025blip3}, and CC12M~\cite{changpinyo2021conceptual}) shown in~\cref{tab:encoder_rank_supp}, and the phenomenon is consistent with~\cref{tab:encoder_rank}.

\begin{table}[ht]
\centering
\small
\caption{\textbf{\footnotesize Effective dimensionality analysis on ImageNet-21k~\cite{ridnik2021imagenetk}, BLIP3o~\cite{chen2025blip3}, and CC12M~\cite{changpinyo2021conceptual}.}}
\setlength{\tabcolsep}{5pt}
\resizebox{0.9\linewidth}{!}{\begin{tabular}{l c ccc ccc ccc}
\toprule
\multirow{2}{*}{\textbf{Tokenizer/Encoder}} & \multirow{2}{*}{\textbf{Full dim}}
& \multicolumn{3}{c}{\textbf{ImageNet-21k}}
& \multicolumn{3}{c}{\textbf{BLIP3o}}
& \multicolumn{3}{c}{\textbf{CC12M}} \\
\cmidrule(lr){3-5} \cmidrule(lr){6-8} \cmidrule(lr){9-11}
& & $R_{90}$ & $R_{95}$ & $R_{99}$ & $R_{90}$ & $R_{95}$ & $R_{99}$ & $R_{90}$ & $R_{95}$ & $R_{99}$ \\
\midrule
DINOv1-B~\cite{caron2021emerging}            & 768  & 499 & 594 & 701 & 483 & 582 & 697 & 481 & 581 & 696 \\
\cdashlinelr{1-11}
DINOv2-B-RAE~\cite{oquab2023dinov2,zheng2025diffusion}    & 768  & 505 & 610 & 726 & 491 & 601 & 723 & 494 & 602 & 724 \\
SigLIP2-B-RAE~\cite{tschannen2025siglip,zheng2025diffusion}   & 768  & 466 & 583 & 716 & 470 & 585 & 717 & 488 & 598 & 721 \\
MAE-B-RAE~\cite{he2022masked,zheng2025diffusion}      & 768  &  92 & 228 & 562 &  77 & 200 & 536 &  74 & 196 & 533 \\
MAE-B-IG-3B*~\cite{singh2023effectiveness}       & 768  & 185 & 335 & 588 & 204 & 356 & 600 & 201 & 355 & 600 \\
\cdashlinelr{1-11}
DINOv2-L~\cite{oquab2023dinov2}           & 1024 & 671 & 811 & 966 & 659 & 804 & 964 & 660 & 804 & 963 \\
Finetuned DINOv2-L                          & 1024 & 124 & 293 & 720 & 114 & 276 & 707 & 121 & 287 & 716 \\
\bottomrule
\end{tabular}}
\label{tab:encoder_rank_supp}
\end{table}

\section{Understanding Performance of Finetuned Tokenizer}
Following~\cite{chen2025aligning}, we use linear probing on ImageNet-1k~\cite{deng2009imagenet} to test the understanding capability of finetuned tokenizer. Original DINOv2-L~\cite{oquab2023dinov2} achieves 84.5$\%$ accuracy, while ours achieves 63.9$\%$. Conversely, our finetuned DINOv2 could achieve a PSNR of 29.12, while PSNR of original DINOv2~\cite{oquab2023dinov2} is 17.34. How to preserve good understanding capability and improve reconstruction performance for a single tokenizer is future research.

\section{Visual Understanding of Frozen VLM Backbone}

To verify that the frozen VLM backbone (Qwen3-VL-2B-Instruct~\cite{bai2025qwen3}) provides a sufficiently strong foundation for text-conditioned generation, we evaluate its visual reasoning and understanding capabilities for reference. We assess its performance on seven representative multimodal benchmarks, including MMMU~\cite{yue2024mmmu}, MMMU-Pro~\cite{yue2025mmmu}, MathVista~\cite{lu2023mathvista}, MathVerse~\cite{zhang2024mathverse}, MathVision~\cite{wang2024measuring}, MMStar~\cite{chen2024we}, and Charxiv-RQ~\cite{wang2024charxiv}. The results are shown in~\cref{tab:understanding}.

\begin{table}[h!]
\centering
\small
\caption{Performance of frozen VLM backbone across seven visual reasoning and understanding benchmarks.}
\resizebox{\textwidth}{!}{
\begin{tabular}{l c c c c c c c}
\toprule
 & MMMU & MMMU-Pro & MathVista & MathVerse & MathVision & MMStar & Charxiv-RQ\\
\toprule  
Frozen VLM Backbone & 52.9 & 37.2 & 61.9 & 53.2 & 30.9 & 58.5 & 25.8\\  
\bottomrule
\end{tabular}
}
\label{tab:understanding}
\end{table}

\section{Extended Derivations for High-Dimensional Flow Matching}
\label{sec:supp_flow_matching}

\paragraph{Notation}
In Sec.~3.2 of the main paper, the learned predictors $\boldsymbol{v}_{\theta}$ and $\boldsymbol{x}_{\theta}$ are trained to approximate the corresponding Bayes-optimal predictors under squared loss. More precisely, we define these optimal high-dimensional predictors as
\begin{equation}
    \boldsymbol{v}^{h,*}(\boldsymbol{z}_t^h,t)
    := \mathbb{E}[\boldsymbol{\epsilon}_h - \boldsymbol{x}_0 \mid \boldsymbol{z}_t^h],
    \qquad
    \boldsymbol{x}^{h,*}(\boldsymbol{z}_t^h,t)
    := \mathbb{E}[\boldsymbol{x}_0 \mid \boldsymbol{z}_t^h],
\end{equation}
and similarly in the low-dimensional space,
\begin{equation}
    \boldsymbol{v}^{l,*}(\boldsymbol{z}_t^l,t)
    := \mathbb{E}[\boldsymbol{\epsilon}_l - \boldsymbol{c}_0 \mid \boldsymbol{z}_t^l],
    \qquad
    \boldsymbol{x}^{l,*}(\boldsymbol{z}_t^l,t)
    := \mathbb{E}[\boldsymbol{c}_0 \mid \boldsymbol{z}_t^l].
\end{equation}

\paragraph{Setup}
Let the observed high-dimensional feature be $\boldsymbol{x}_{0} \in \mathbb{R}^{h}$ and the true latent variable be $\boldsymbol{c}_{0} \in \mathbb{R}^{l}$. For simplicity, we assume the relationship is linear: $\boldsymbol{x}_{0} = Q\boldsymbol{c}_{0}$, where the columns of $Q \in \mathbb{R}^{h \times l}$ form an orthonormal basis for the subspace (i.e., $Q^\top Q = I_l$). We denote the orthogonal projectors $P_Q = QQ^\top$ and $P_\perp = I_h - QQ^\top$.

\paragraph{Orthogonal decomposition of the forward process}
In standard flow matching, the forward process constructs the noisy state $\boldsymbol{z}_{t}^{h} = (1-t)\boldsymbol{x}_{0} + t\boldsymbol{\epsilon}_{h}$, where $\boldsymbol{\epsilon}_{h} \sim \mathcal{N}(0, I_h)$ and $t \in [0, 1]$ ($t{=}0$ corresponds to clean data, $t{=}1$ to pure Gaussian noise).

Because $\boldsymbol{x}_0$ is assumed to lie in the column space of $Q$, we can decompose the isotropic Gaussian noise $\boldsymbol{\epsilon}_{h}$ into two orthogonal components: a component parallel to the data subspace, $\boldsymbol{\epsilon}_{l} = Q^\top \boldsymbol{\epsilon}_{h} \sim \mathcal{N}(0, I_l)$, and a component normal to it, $\boldsymbol{\epsilon}_{\perp} = P_\perp\boldsymbol{\epsilon}_{h}$. These two components are independent, since $\mathrm{Cov}(\boldsymbol{\epsilon}_l, \boldsymbol{\epsilon}_\perp) = Q^\top P_\perp = Q^\top - Q^\top QQ^\top = 0$ and both are Gaussian. Moreover, $\boldsymbol{c}_0$ (drawn from the data distribution) is independent of $\boldsymbol{\epsilon}_h$, and hence independent of both $\boldsymbol{\epsilon}_l$ and $\boldsymbol{\epsilon}_\perp$.

Substituting into the forward process:
\begin{equation}
    \boldsymbol{z}_{t}^{h} = Q\underbrace{((1-t)\boldsymbol{c}_{0} + t\boldsymbol{\epsilon}_{l})}_{\boldsymbol{z}_{t}^{l}} + \; t\boldsymbol{\epsilon}_{\perp},
\end{equation}
where $\boldsymbol{z}_{t}^{l} \in \mathbb{R}^l$ is the manifold component. Crucially, $\boldsymbol{z}_t^l$ is a function of $(\boldsymbol{c}_0, \boldsymbol{\epsilon}_l)$ only, and is therefore independent of $\boldsymbol{\epsilon}_\perp$. Applying $P_\perp$ to both sides isolates the normal component deterministically for $t > 0$:
\begin{equation}
    P_\perp\boldsymbol{z}_{t}^{h} = t\boldsymbol{\epsilon}_{\perp}, \quad \text{i.e.,} \quad \boldsymbol{\epsilon}_{\perp} = \tfrac{1}{t}P_\perp\boldsymbol{z}_{t}^{h}.
    \label{eq:supp_eps_perp_deterministic}
\end{equation}
This means $\boldsymbol{\epsilon}_\perp$ is fully determined by $\boldsymbol{z}_t^h$---no statistical estimation is required to recover it. We also note that $Q^\top\boldsymbol{z}_t^h = \boldsymbol{z}_t^l$ (since $Q^\top \boldsymbol{\epsilon}_\perp = Q^\top P_\perp \boldsymbol{\epsilon}_h = 0$), so the model can recover the manifold state from $\boldsymbol{z}_t^h$ via projection.

\begin{figure}[!t]
\centering

\includegraphics[width=0.8\linewidth]{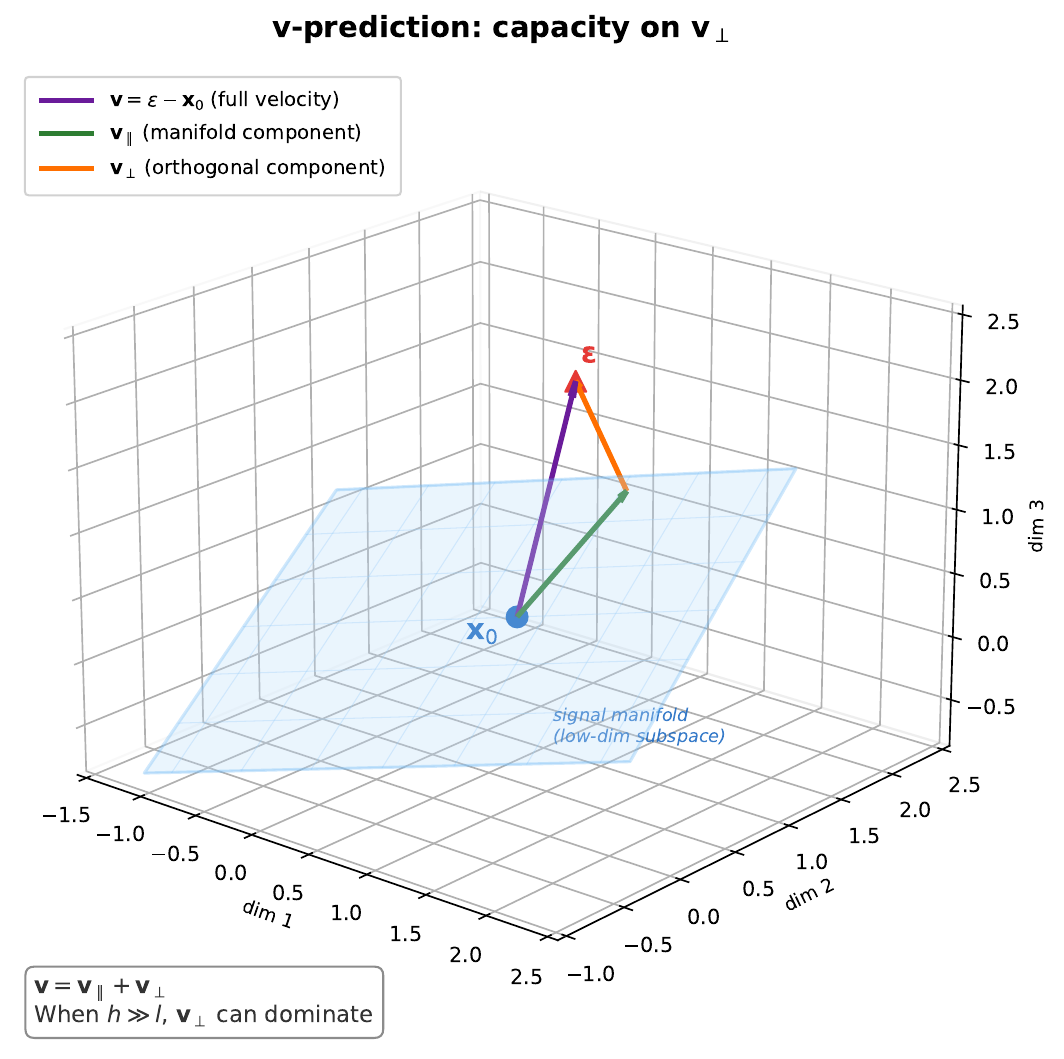}

\caption{\textbf{Decomposition of the velocity target under the subspace model.} 
The full velocity $\boldsymbol{v} = \boldsymbol{\epsilon} - \boldsymbol{x}_0$ 
(purple) decomposes into a manifold-aligned component $\boldsymbol{v}_{\parallel}$ 
(green) and an orthogonal component $\boldsymbol{v}_{\perp}$ (orange). 
When $h \gg l$, $\boldsymbol{v}_{\perp}$ can dominate the target energy, 
forcing the model to spend capacity on an uninformative direction. 
In contrast, the $\boldsymbol{x}_0$-prediction target 
$\boldsymbol{x}_0 = Q\boldsymbol{c}_0$ lies entirely on the signal manifold, 
bypassing $\boldsymbol{v}_{\perp}$ altogether.}

\label{fig:velocity_viz}
\end{figure}

\paragraph{The inefficiency of high-dimensional velocity ($\boldsymbol{v}$) prediction}
The velocity target is $\boldsymbol{v} = \boldsymbol{\epsilon}_{h} - \boldsymbol{x}_{0}$. Using the decomposition $\boldsymbol{\epsilon}_h = Q\boldsymbol{\epsilon}_l + \boldsymbol{\epsilon}_\perp$ and $\boldsymbol{x}_0 = Q\boldsymbol{c}_0$:
\begin{equation}
    \boldsymbol{v}^{h,*}(\boldsymbol{z}_{t}^{h}, t) = \mathbb{E}[\boldsymbol{v} \mid \boldsymbol{z}_{t}^{h}] = \mathbb{E}[Q(\boldsymbol{\epsilon}_{l} - \boldsymbol{c}_{0}) \mid \boldsymbol{z}_{t}^{h}] + \mathbb{E}[\boldsymbol{\epsilon}_{\perp} \mid \boldsymbol{z}_{t}^{h}].
\end{equation}
By~\cref{eq:supp_eps_perp_deterministic}, the second term equals $\frac{1}{t}P_\perp\boldsymbol{z}_t^h$. For the first term, since $\boldsymbol{\epsilon}_l - \boldsymbol{c}_0$ depends only on $(\boldsymbol{c}_0, \boldsymbol{\epsilon}_l)$, which are jointly independent of $\boldsymbol{\epsilon}_\perp$, conditioning on $\boldsymbol{z}_t^h = (Q\boldsymbol{z}_t^l, t\boldsymbol{\epsilon}_\perp)$ reduces to conditioning on $\boldsymbol{z}_t^l$ alone. Therefore:
\begin{equation}
    \boldsymbol{v}^{h,*}(\boldsymbol{z}_{t}^{h}, t) = Q\boldsymbol{v}^{l,*}(Q^\top \boldsymbol{z}_{t}^{h}, t) + \frac{1}{t}P_\perp\boldsymbol{z}_{t}^{h},
\label{eq:supp_velocity}
\end{equation}
where $\boldsymbol{v}^{l,*}(\boldsymbol{z}_t^l, t) = \mathbb{E}[\boldsymbol{\epsilon}_l - \boldsymbol{c}_0 \mid \boldsymbol{z}_t^l]$ is the optimal velocity predictor in the $l$-dimensional latent space.

This reveals a critical inefficiency: the second term is a deterministic, $(h-l)$-dimensional linear function of the input that lies orthogonal to the data subspace and carries no additional information about the clean latent $\boldsymbol{c}_0$ beyond what is already contained in $\boldsymbol{z}_t^l$, yet the model must still allocate capacity to represent it. In terms of target norm, $\mathbb{E}[\|\boldsymbol{\epsilon}_\perp\|^2] = \mathrm{tr}(P_\perp) = h - l$, while the low-dimensional velocity target has expected squared norm $l + \mathbb{E}[\|\boldsymbol{c}_0\|^2]$. Using $R_{90}$ as an empirical proxy for the intrinsic dimension, the finetuned DINOv2-L features ($h=1024$, $R_{90}=129$) suggest that a large fraction of the ambient space lies outside the effective signal subspace (see~\cref{fig:velocity_viz}).

\paragraph{The efficiency of $\boldsymbol{x}_0$-prediction}
To bypass modeling the orthogonal noise, we consider direct $\boldsymbol{x}_{0}$-prediction, which is also inspired by~\cite{li2025back}:
\begin{equation}
    \boldsymbol{x}^{h,*}(\boldsymbol{z}_{t}^{h}, t) = \mathbb{E}[\boldsymbol{x}_{0} \mid \boldsymbol{z}_{t}^{h}] = \mathbb{E}[Q\boldsymbol{c}_{0} \mid Q\boldsymbol{z}_{t}^{l}, t\boldsymbol{\epsilon}_{\perp}].
\label{eq:supp_x0_pred}
\end{equation}
Since $(\boldsymbol{c}_0, \boldsymbol{\epsilon}_l)$ are jointly independent of $\boldsymbol{\epsilon}_\perp$, and $\boldsymbol{z}_t^l$ is a function of $(\boldsymbol{c}_0, \boldsymbol{\epsilon}_l)$ alone, conditioning on $\boldsymbol{z}_t^l$ cannot introduce dependence on $\boldsymbol{\epsilon}_\perp$, i.e., $\boldsymbol{c}_0 \perp\!\!\!\perp \boldsymbol{\epsilon}_\perp \mid \boldsymbol{z}_t^l$. Therefore, $t\boldsymbol{\epsilon}_\perp$ provides zero additional information about $\boldsymbol{c}_0$ beyond what $\boldsymbol{z}_t^l$ already contains, and the conditional expectation simplifies to:
\begin{equation}
    \boldsymbol{x}^{h,*}(\boldsymbol{z}_{t}^{h}, t) = \mathbb{E}[Q\boldsymbol{c}_{0} \mid Q\boldsymbol{z}_{t}^{l}] = Q\mathbb{E}[\boldsymbol{c}_{0} \mid \boldsymbol{z}_{t}^{l}] = Q\boldsymbol{x}^{l,*}(\boldsymbol{z}_{t}^{l}, t).
\label{eq:supp_x0_simplified}
\end{equation}
\cref{eq:supp_x0_simplified} shows that, under the subspace model, the Bayes-optimal $\boldsymbol{x}_0$-predictor outputs vectors in
$\mathrm{col}(Q)$, with no orthogonal noise component in the clean-data target. This provides a parameterization-based explanation consistent with the optimization improvements observed empirically.

\end{document}